\documentclass{llncs}

\usepackage{xcolor}
\usepackage[compatibility=false]{caption}
\usepackage{subcaption}
\usepackage{graphicx}
\usepackage{amsmath}
\usepackage{amssymb}
\usepackage{cleveref}
\usepackage{multirow}
\usepackage{booktabs}
\usepackage{marvosym}

\crefformat{section}{\S#2#1#3} 
\crefformat{subsection}{\S#2#1#3}
\crefformat{subsubsection}{\S#2#1#3}

\begin{document}

\title{On the Effect of Inter-observer Variability for a Reliable Estimation of Uncertainty of Medical Image Segmentation}
\titlerunning{Inter-observer Variability and Uncertainty}  

\author{Alain Jungo\inst{1}\textsuperscript{(\Letter)} \and Raphael Meier\inst{2} \and Ekin Ermis\inst{3} \and Marcela Blatti-Moreno\inst{3} \and Evelyn Herrmann\inst{3} \and Roland Wiest\inst{2} \and Mauricio Reyes\inst{1}}
\authorrunning{Alain Jungo et al.} 
\institute{Institute for Surgical Technologies and Biomechanics, \\
University of Bern, Switzerland\and SCAN, Institute for Diagnostic and Interventional Neuroradiology,\and University Clinic for Radio-oncology,\\
Inselspital, Bern University Hospital, University of Bern, Switzerland\\
\email{alain.jungo@istb.unibe.ch}} 


\maketitle              

\begin{abstract}
Uncertainty estimation methods are expected to improve the understanding and quality of computer-assisted methods used in medical applications (e.g., neurosurgical interventions, radiotherapy planning), where automated medical image segmentation is crucial. In supervised machine learning, a common practice to generate ground truth label data is to merge observer annotations. However, as many medical image tasks show a high inter-observer variability resulting from factors such as image quality, different levels of user expertise and domain knowledge, little is known as to how inter-observer variability and commonly used fusion methods affect the estimation of uncertainty of automated image segmentation. In this paper we analyze the effect of common image label fusion techniques on uncertainty estimation, and propose to learn the uncertainty among observers. The results highlight the negative effect of fusion methods applied in deep learning, to obtain reliable estimates of segmentation uncertainty. Additionally, we show that the learned observers' uncertainty can be combined with current standard Monte Carlo dropout Bayesian neural networks to characterize uncertainty of model's parameters. 

\end{abstract}
\section{Introduction}
The performance of medical image segmentation has increased with the advances in supervised machine learning and is reported to achieve close to human performance for specific tasks \cite{Meier2016}. Despite the success of deep learning and its merit in recent state-of-the-art methods \cite{Shen2017}, modern systems still lack in robustness and yield unexpected errors which hinders the adoption of such systems in medical applications. Uncertainty estimates of computer's results can help to foster understanding and trustworthiness of the underlying deep learning models. Various works have been proposed to produce uncertainty estimates in neural networks \cite{blundell2015weight,Gal2016Bayesian}. The Bayesian approach through Monte Carlo dropout proposed by Gal and Ghahramani \cite{Gal2016Bayesian} is probably the most popular due to its simple realization. Most methods built on Bayesian approaches stem from computer vision applications whereon ground truth definition has low inter-observer variability. However, calculation of segmentation uncertainty in medical images is particularly difficult, as the image content and quality can vary (e.g., image resolution, patient motion, partial volume effect), and often times medical images only partially describe the anatomy or (patho)physiology of interest. This can lead to a large inter-observer variability that is exacerbated by clinical domain-knowledge required to manually segment medical images.  To deal with inter-observer variability in medical image segmentation, supervised learning approaches are typically trained using ground truth generated by common fusion techniques (e.g., majority voting, STAPLE \cite{warfield2002validation}). However, as inter-observer variability reflects the disagreement among experts, we postulate that a supervised learning approach needs to likewise reflect experts disagreement when providing uncertainty estimates on new unseen cases. Little is known as to how inter-observer variability and commonly used fusion methods affect the estimation of image segmentation uncertainty. We hypothesize that inter-observer variability needs to be taken into account when learning models aiming at producing reliable estimations of segmentation uncertainty. 

To this end, in this paper we analyze the effect of common image label fusion techniques on uncertainty estimation, and propose to learn the uncertainty among observers. Additionally, we show that the learned observers' uncertainty can be combined with current standard Monte Carlo dropout Bayesian neural networks to characterize uncertainty of model's parameters.
Due to the absence of a real ground truth in medical images we first analyze the effect on a synthetic dataset that simulates inter-observer variability.
In a final experiment, we analyze the behavior on a clinical post-operative brain tumor cavity dataset with multiple observer annotations.

\section{Uncertainty Estimation in Deep Learning}
This section introduces two types of uncertainty considered below: uncertainty linked to inter-observer variability, and the intrinsic model's uncertainty linked to the difficulty of a given model to make a prediction. 

\subsection{Uncertainty linked to inter-observer variability}\label{subsec:rater}

\begin{figure}[!t]
  \centering
  \includegraphics[width=0.85\textwidth]{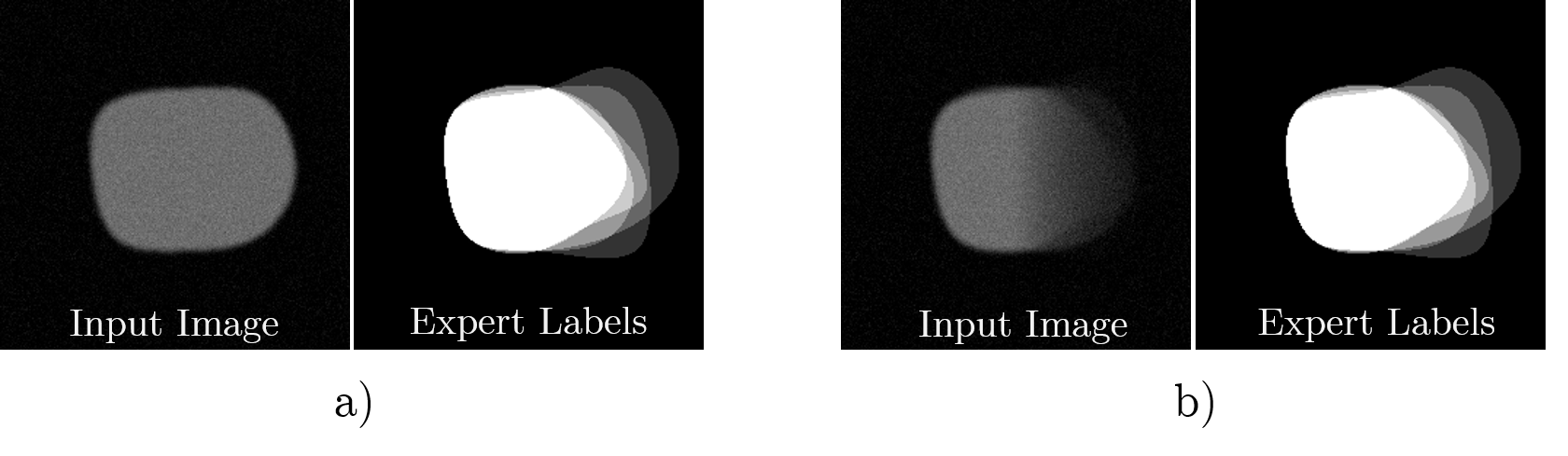}
  \caption{Synthetic analysis using two simulated image-label scenarios. (a) low entropy of the input image and inter-observer variability, (b) high entropy of the input image and inter-observer variability.}
  \label{fig:situations}
\end{figure}

We analyze uncertainty linked to inter-observer variability through simulated scenarios including inter-observer variability and two different levels of image entropy. These are: (i) inter-observer variability and low entropy of the input image (i.e. crisp or sharp image edges), and (ii) inter-observer variability and high entropy of the input image (i.e. diffuse image edges). While case (ii) better reflects the reality in medical applications, case (i) was created to test whether the image content (in terms of difficulty of the segmentation task) affects the estimation of uncertainty. Following the initial postulate, we are interested to analyze the model's capability to learn the inter-observer variability into the estimation of segmentation uncertainty regardless of the image content. Figure~\ref{fig:situations} illustrates these configurations. 

\subsection{Uncertainty linked to model's parameters}
Parameter uncertainty can be modeled by Bayesian neural networks \cite{MacKay:1992:PBF:148147.148165,neal2012bayesian} with distributions for the model weights. As presented by Gal and Ghahramani \cite{Gal2016Bayesian}, dropout regularization can be interpreted as an approximation for Bayesian inference over the weights of the network. If applied at test time, dropout produces randomly sampled networks, which can be viewed as Monte Carlo samples over the posterior distribution of the model weights. 
Be $I$ an input image that leads to a predicted class $y_i \in \mathcal{C}$ at pixel $i$, where $\mathcal{C}$ is the set of classes. Then the approximative class probability resulting from $T$ Monte Carlo samples is $ p(y_i=c \mid I) \approx \frac{1}{T}\sum_{t=1}^{T}{p(y_{i,t}=c \mid I, W^t)}$ with sampled weights $W^t$. The uncertainty can be computed by the predictive entropy $\mathit{H} \approx -\sum_{c \in \mathcal{C}}{p(y_i=c \mid I)\log p(y_i=c \mid I)}$.
With increasing dataset size the model's parameter uncertainty decreases \cite{kendall2017uncertainties}. This makes it well-suited for the use in medical images, where training datasets are typically small.
Since computing uncertainty estimations via Monte Carlo dropout does not pose any restrictions on the learning procedure, the uncertainty linked to the inter-observer variability can be combined with the model's parameter uncertainty.

\section{Experimental Setup}
\subsection{Deep Learning Architecture}
For both experiments, we used a U-net-based \cite{Ronneberger2015} architecture. We chose this architecture because of its popularity and its vast use in the medical imaging domain. We modify the standard architecture by adding a dropout layer ($p=0.2$) after each convolution layer \cite{kendall2017uncertainties}. For all experiments we use $T=20$ Monte Carlo samples.

\subsection{Experiment 1: Synthetic}
In this experiment, we aim at examining the impact of the fusion method (or absence of) on the uncertainty estimation. We analyzed the following approaches: (a) no fusion (i.e. all labels used during training), (b) majority vote, (c) STAPLE \cite{warfield2002validation}, (d) intersection (to simulate a strict expert agreement) and (e) union of all observers (all experts' results are merged).

We produced a synthetic dataset to circumvent the absence of a multi-observer dataset with known underlying ground truth. The dataset aims at mimicking the situations described in \cref{subsec:rater} without introducing additional complexity. A synthetic sample of the dataset is created in four steps, as follows:
\vspace{0.5em}

\textit{Ground truth generation:} Eight perimeter points, initially equidistantly lying on a circle, are randomly perturbed with respect to the circle's center (angles: $\pm15^\circ$, distance factor: $[0.75, 1.5]$) and interpolated with a B-spline model. 

\textit{Low-entropy (i.e. unperturbed) images:} Input images $I$ were derived from the ground truth by: (a) varying the maximum value (initially 255) randomly $[30,255]$, (b) adding a Gaussian blur with random sigma $[2,8]$, (c) adding Gaussian noise (factor: $0.15(\max(I) - \min(I)$). See input image of Fig.~\ref{fig:situations}(a).

\textit{Observer annotations:} Perturbations to the simulated expert annotations were conducted by randomly perturbing the three rightmost perimeter points of the ground truth (angles: $\pm10^\circ$, distance: $\pm0.4\cdot d$ with $d$ the distance to the center). 

\textit{High-entropy (i.e. perturbed) image:} Observer annotations are first summed up to the ground truth with random intensities $[50, 255]$, followed by an intensity normalization. Afterwards, maximum intensities are randomized ($[30,255]$); an intensity gradient is added (random exponential decay $[0.5, 6.5]$ towards the right part of the image, and Gaussian blur (random sigma $[2,8]$) and noise (factor: $0.15(\max(I) - \min(I)$) are introduced. See input image of Fig.~\ref{fig:situations}(b).
\vspace{0.5em}

The dataset of synthetic images consists of 100 samples, each containing a ground truth, five observer annotations, and perturbed and unperturbed grayscale images.

\subsubsection{Implementation details.}
Due to the low complexity of the task for the synthetic data, the U-net model consists only of two pooling/upsampling steps with an initial filter size of 16. For all fusion methods (and the absence of fusion) a network was trained for 100 epochs, with the last model being selected. In the absence of fusion, the observer annotations are sampled randomly. Adam \cite{KingmaB15} optimizer with a learning rate of $10^{-3}$ was used.

\subsection{Experiment 2: Brain Tumor Cavity}
In this experiment, we aim at validating the findings of the synthetic experiment on clinical data. We compared the uncertainty obtained by training without fusion and chose majority vote as fusion method yielding best segmentation performance.
Since the underlying ground truth is unknown, a qualitative evaluation of the segmentation was performed. 

\subsubsection{MRI patient data.}
The clinical dataset consists of 30 post-operative brain tumor magnetic resonance images, with isotropic voxel size ($1\times1\times1 mm$) acquired in the four standard sequences (T1-weighted (T1), T1-weighted post-contrast (T1c), T2-weighted (T2) and Fluid-attenuated inversion-recovery (FLAIR)), which are used to evaluate post-operative status of glioblastoma patients. The binary label maps delineate the cavity after tumor resection, and it is used for radiotherapy planning. The dataset contains annotations of three clinical radiation oncology experts with different levels of expertise (two years, four years, and over six years of clinical experience). This dataset is particularly interesting as post-operative resection cavities are ill-defined due to the presence of blood products producing pseudo-image gradients, CSF infiltration and air pockets. 

\subsubsection{Implementation details.}
To adapt to the much more complex task of segmenting post-operative brain tumor resection cavities, we chose five pooling/upsampling steps and a initial filter size to 48. We used a two-dimensional input of the network and applied it on the axial slices of the brain volumes. The networks were trained for 35 epochs with selection of the last model. The optimizer is Adam with learning rate $10^{-3}$. Due to the small dataset size, a six-fold cross-validation was performed. 

\subsection{Evaluation Metrics}
As postulated, we seek reliable estimations of uncertainty that reflect expert disagreement as a result of the complexity of the task and different levels of expertise. As part of the Asimolar set of principles in A.I, this is known as \textit{capability caution} on the upper limits of performance for systems learning from experts. To assess this, we assessed how fusion methods (or absence of) affect uncertainty in regions where expert disagreement is observed. We quantify this via  ${WME} = \frac{1}{N}\sum_{i = 1}^{N}{\mathit{\hat{H}}_i\cdot \mathit{H}_i}$, which is the weighted mean of the predictive entropy $\mathit{H}$ over $N$ pixels, and with $\mathit{\hat{H}}$ corresponding to the entropies of the expert disagreement.
In order to capture the overall uncertainty produced by a model, we also evaluated the mean predictive entropy. This allows us to detect models yielding a high uncertainty but not reflecting the disagreement among the experts. Additionally, the Dice coefficient was used to assess segmentation performance.

\section{Results \& Discussion}

\subsection{Experiment 1: Synthetic}

The weighted mean entropy (WME), the mean entropy (ME), and the Dice coefficient were computed in relation to the known ground truth. Quantitative results are shown in Table~\ref{tab:qunatitative_synthetic} and Fig.~\ref{fig:boxplots}(a). The results particularly highlight the simultaneous increase of WME and ME for the model trained without any label fusion, as compared to the other models. This suggests that the uncertainty derived by a model trained with all labels (i.e. no fusion) better describes expert disagreement. On the contrary, training with intersection or union of labels reduces the reliability of the estimated uncertainty.
In terms of segmentation performance, it is observed that training with all labeled information (i.e. no fusion) performs as well as those trained with either majority voting or STAPLE. This result suggests that a more reliable uncertainty does not come with a reduced segmentation accuracy, as it could have been expected when training models with non-fused label data. 

\begin{figure}[!ht]
  \centering
  \includegraphics[width=1\textwidth]{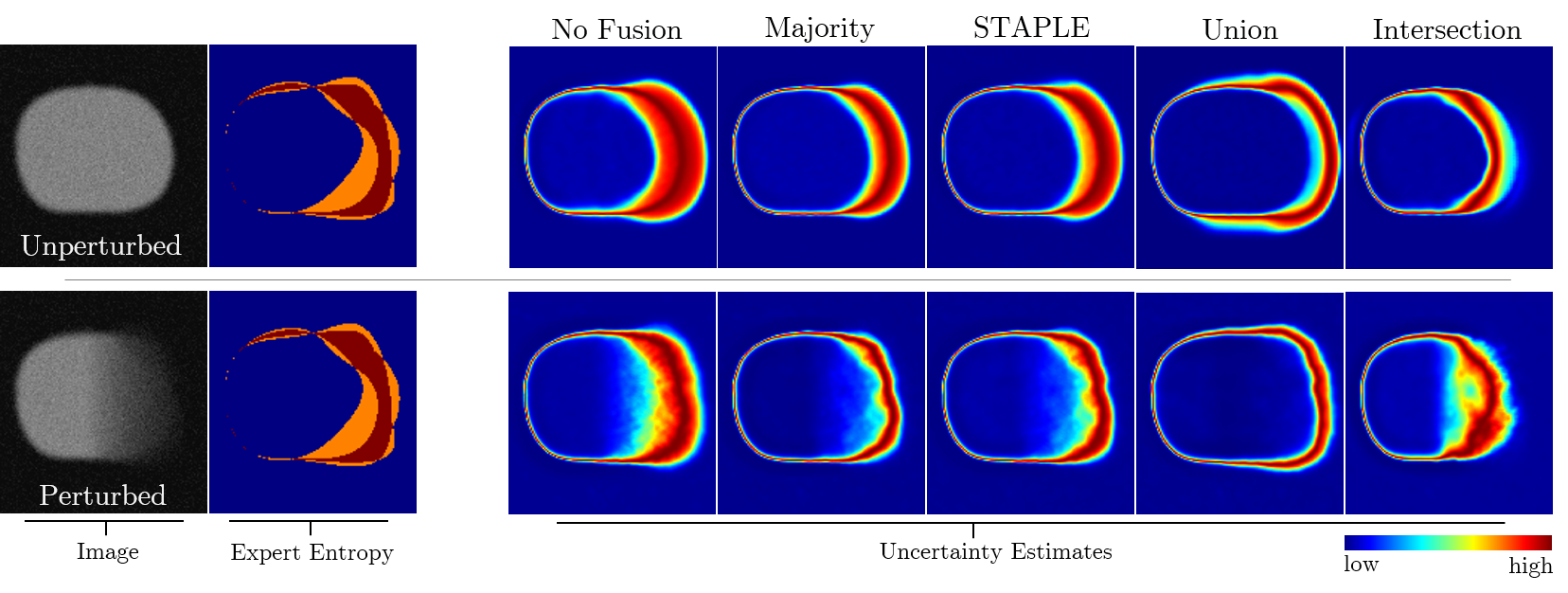}
  \caption{Uncertainty estimations obtain from models trained on differently fused labels on the synthetic dataset. Top and bottom row correspond to training with unperturbed and perturbed input images, respectively. Columns correspond to the fusion method used.}
  \label{fig:synthetic_uncertainty}
\end{figure}

Figure~\ref{fig:synthetic_uncertainty} presents qualitative results, showing that models trained with fused labels tend to underestimate the uncertainty with respect to the reference expert variability (second column of Fig~\ref{fig:synthetic_uncertainty}). Conversely, the model trained without any fusion better resembles the reference expert variability.
Results on the unperturbed images (top row of Fig.~\ref{fig:synthetic_uncertainty}) show that despite of the clear edge information of the input image, the uncertainty estimates reflect the underlying expert disagreement. This result verifies our postulate that a model can learn inter-observer variability regardless of the image content.
\begin{table}
  \centering
  \caption{Quantitative results of the synthetic experiment. The fusion methods are compared on weighted mean entropy (WME), mean entropy (ME) and Dice coefficient. \textit{U} and \textit{P} stand for \textit{unperturbed} and \textit{perturbed} and describe the state of the input image.}
  \label{tab:qunatitative_synthetic}
  \scriptsize
  \begin{tabular}{c@{\hskip 1em} c@{\hskip 2em} c@{\hskip 1em} c@{\hskip 1em} c@{\hskip 1em} c@{\hskip 1em} c}
  \toprule
   & & \textbf{No fusion} & \textbf{Majority} & \textbf{STAPLE} & \textbf{Union} & \textbf{Intersection}\\
  \midrule
  \multirow{2}{*}{\textbf{WME}}
   & \textit{U} & \textbf{.68}$\pm$.08 & .54$\pm$.10 & .59$\pm$.09 & .29$\pm$.13 & .30$\pm$.12 \\
   & \textit{P} & \textbf{.67}$\pm$.07 & .47$\pm$.09 & .56$\pm$.08 & .30$\pm$.08 & .20$\pm$.08 \vspace{.3em}\\   
  \multirow{2}{*}{\textbf{ME}}
   & \textit{U} & \textbf{.12}$\pm$.03 & .09$\pm$.02 & .10$\pm$.02 & .08$\pm$.02 & .08$\pm$.02 \\
   & \textit{P} & \textbf{.12}$\pm$.03 & .08$\pm$.02 & .09$\pm$.02 & .08$\pm$.02 & .09$\pm$.03 \vspace{0.3em}\\
  \multirow{2}{*}{\textbf{Dice}}
   & \textit{U} & \textbf{.99}$\pm$.01 & \textbf{.99}$\pm$.01 & .98$\pm$.01 & .90$\pm$.02 & .89$\pm$.03 \\
   & \textit{P} & \textbf{.96}$\pm$.02 & \textbf{.96}$\pm$.02 & \textbf{.96}$\pm$.02 & .92$\pm$.03 & .88$\pm$.04\\
  \bottomrule
  \end{tabular}
\end{table}

\subsection{Experiment 2: Brain Tumor Cavity}

\begin{figure}[!ht]
  \centering
  \includegraphics[width=1\textwidth]{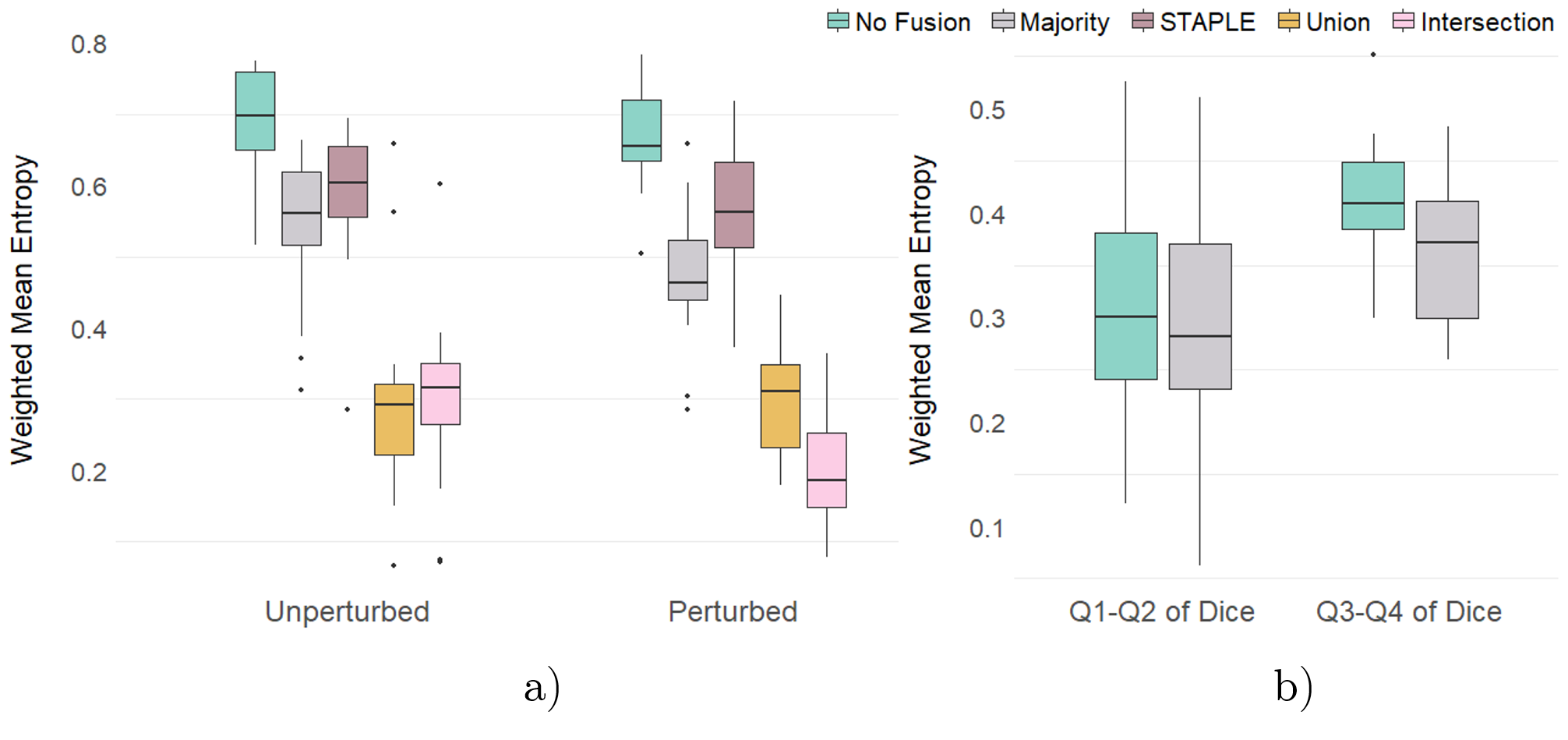}
  \caption{Quantitative results of the weighted mean entropy (WME) metric. (a) results obtained for fusion methods on perturbed and unperturbed images of the synthetic dataset, (b) cavity dataset results separated by Dice performance (Q1-Q2: below median, Q3-Q4: above median).}
  \label{fig:boxplots}
\end{figure}

Figure~\ref{fig:boxplots}(b) illustrates the obtained variability of WME on the 30 cross-validated cavity images. Results were divided by segmentation performance in two groups, separated by the median Dice. For underperforming segmentation results (below median), results show no benefit in employing all label data for training and estimating uncertainty. Conversely, for segmentation results where the Dice was equal or larger than the median Dice, a benefit on using all label data was observed. This result suggests the existence of a link between segmentation performance and reliability of uncertainty estimation.

\begin{figure}[!ht]
  \centering
  \includegraphics[width=0.85\textwidth]{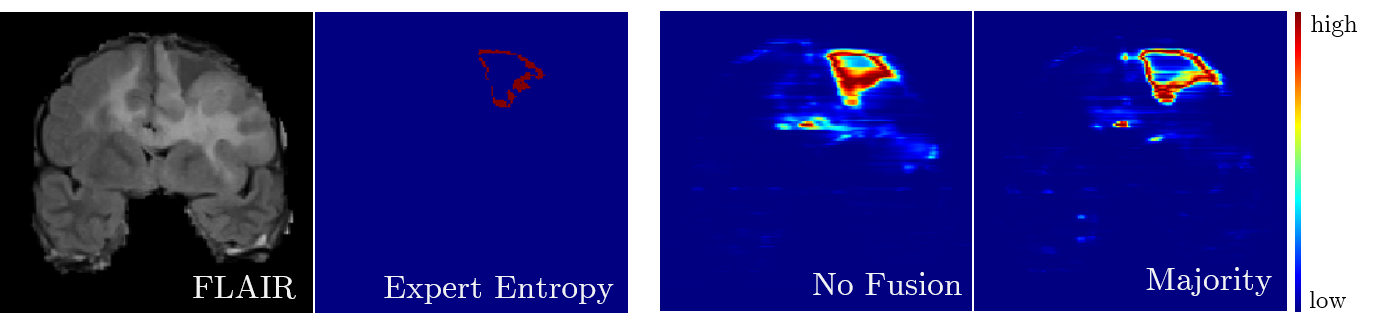}
  \caption{Exemplary uncertainty estimations on the cavity dataset in comparison to the expert entropy.}
  \label{fig:cavity_uncertainty}
\end{figure}

Figure~\ref{fig:cavity_uncertainty} presents a qualitative result. It shows that the model trained on all labels is able to produce reliable uncertainty in regions of highest expert disagreement (right cavity side).

\vspace{1em}
The experiments on the synthetic and clinical dataset reveal that uncertainty estimations is linked to inter-observer variability, and that reliable uncertainty (i.e. reflecting expert disagreement) may be learned by avoiding label fusion in the training data. This is of high relevance in systems where for example, uncertainty estimations are used by experts to monitor and correct computer-generated results. 
In addition, we observed a link between segmentation performance and reliability of segmentation uncertainty estimates.

\section{Conclusion}
In this paper, we analyzed the impact of fusion methods on the uncertainty estimation. Experiments were performed on a synthetic multi-observer dataset and a clinical dataset. First evidence verifies the link between uncertainty estimations from trained deep learning models and inter-observer variability, which is inherent of medical image applications. We conclude that the benefit of using fusion methods for reliable segmentation uncertainty estimations is conditioned to the performance of the underlying segmentation accuracy, and hence it needs to be assessed when considering fusion methods for ground truth generation.

\subsubsection*{Acknowledgments.} This work was supported by the Swiss National Foundation by grant number 169607.

%
%
\bibliographystyle{splncs03}
\bibliography{references}

\end{document}